\documentclass[11pt,leqno]{article}

\usepackage{amsmath,amssymb,amsfonts,amsthm,bbm,mathrsfs,verbatim} 
\usepackage{algorithm}
\usepackage[noend]{algpseudocode}
\usepackage{setspace}
\usepackage{authblk}
\usepackage{caption}
\usepackage{geometry}
\usepackage{graphics} 
\usepackage{graphicx}  
\usepackage{apacite}
\usepackage{hyperref}
\usepackage[round]{natbib}
\usepackage{fancyhdr} 
\usepackage{color}
\usepackage{bm}

\newtheorem{theorem}{Theorem}[section]

\theoremstyle{definition}
\newtheorem{remark}[theorem]{Remark}
\newtheorem{definition}[theorem]{Definition}
\newtheorem{example}[theorem]{Example}

\def\R{{\mathbb R}}

\def\P{{\mathcal P}}

\newcommand\blue[1]{\textcolor{blue}{#1}}

\title{On clustering uncertain and structured data with Wasserstein barycenters and a geodesic criterion for the number of clusters}

\date{}

\author[a]{\small G. I. Papayiannis\footnote{Corresponding author: Patission Str. 76, Athens, 10434, Greece. E-mail: \blue{\texttt{g.papagiannis@hna.gr}}}}
\author[b]{\small G. N. Domazakis}
\author[c]{\small D. Drivaliaris}
\author[d]{\small S. Koukoulas}
\author[e]{\small A. E. Tsekrekos}
\author[f]{\small A. N. Yannacopoulos}

\affil[a]{\footnotesize Hellenic Naval Academy, Department of Naval Sciences, Section of Mathematics, Mathematical Modeling and Applications Laboratory, Piraeus, GR; Athens University of Economics \& Business, Stochastic Modeling and Applications Laboratory, Athens, GR}
\affil[b]{\footnotesize University of Sussex, Department of Mathematics, Brighton, UK; Athens University of Economics \& Business, Stochastic Modeling and Applications Laboratory, Athens, GR}
\affil[c]{\footnotesize University of the Aegean, Department of Financial and Management Engineering, Chios, GR}
\affil[d]{\footnotesize University of the Aegean, Department of Geography, Mytilene, GR}
\affil[e]{\footnotesize Athens University of Economics \& Business, Department of Accounting and Finance, Athens, GR}
\affil[f]{\footnotesize Athens University of Economics \& Business, Department of Statistics, Stochastic Modeling and Applications Laboratory, Athens, GR}

\begin{document}
\graphicspath{ {figures/} } 

\fancypagestyle{plain}{%
	\fancyhead[R]{\blue{Published in Journal of Statistical Computation and Simulation (\url{https://doi.org/10.1080/00949655.2021.1903463})}}
	\renewcommand{\headrulewidth}{0pt}}

\maketitle

\begin{abstract}
In this work clustering schemes for uncertain and structured data are considered relying on the notion of Wasserstein barycenters, accompanied by appropriate clustering indices based on the intrinsic geometry of the Wasserstein space where the clustering task is performed. Such type of clustering approaches are highly appreciated in many fields where the observational/experimental error is significant (e.g. astronomy, biology, remote sensing, etc.) or the data nature is more complex and the traditional learning algorithms are not applicable or effective to treat them (e.g. network data, interval data, high frequency records, matrix data, etc.). Under this perspective, each observation is identified by an appropriate probability measure and the proposed clustering schemes rely on discrimination criteria that utilize the geometric structure of the space of probability measures through core techniques from the optimal transport theory. The advantages and capabilities of the proposed approach and the geodesic criterion performance are illustrated through a simulation study and the implementation in two real world applications: (a) clustering eurozone countries according to their observed government bond yield curves and (b) classifying the areas of a satellite image to certain land uses categories, a standard task in remote sensing. 
\end{abstract}


\section{Introduction}

The majority of statistical learning methods refer to data which are considered as points on some high-dimensional Euclidean space (e.g. $\R^d$ for $d> 1$), and treat the problem of clustering them into like classes depending on their distance or similarity occurring from their location on the underlying space. Such methods have led to the development of powerful algorithms, as for example the K-means algorithm, which are widely used in statistical learning \cite{hastie2009elements, vapnik2013nature}. On the other hand, the implementation of these algorithms has highlighted the importance of geometrical features such as dimension or the notion of distance employed in the underlying space on the results of the clustering process (\cite{aggarwal2001surprising, bouveyron2007high, de2012minkowski}). 

However, in the new era of data science, the nature of data structures has significantly been enriched and evolved so that in many cases data can no longer be sufficiently represented as elements of some high dimensional Euclidean space like $\R^d$. As a result, traditional methods in statistical learning may not be appropriate for analysing such types of complex data. Examples of such cases can be functional data \cite{chiou2007functional, jacques2014model, ray2006functional}, data sampled at high frequency \cite{goutte1999clustering, liao2005clustering}, interval data \cite{de2006dynamic, de2009partitional, irpino2008dynamic}, spatial data \cite{haggarty2015spatially}, uncertain data \cite{aggarwal2010managing}, matrix-valued data \cite{Tuzel2007HumanDV}, etc. 

In this paper, we provide a framework under which uncertain data or structured data can be effectively separated into groups when projected to an appropriate feature space, and in particular the space of probability measures, which when endowed with an appropriate metric structure, the Wasserstein metric, provides a complete metric framework for learning purposes. After recognizing the appropriate underlying space for the data, clustering schemes in Wasserstein space are discussed, based on its geometry and concept of the barycenter in order to characterize the cluster centroids as well as the concept of geodesic to quantify similarity of observations within a particular cluster and the clusters homogeneity. Moreover, indices that (a) assess the quality of the clustering procedure and (b) provide a very intuitive criterion for the appropriate selection for number of clusters are developed, relying on the intrinsic geometry of the feature space, the space of probability measures. We note that eventhough clustering in the Wasserstein space has been discussed before (see e.g. Section \ref{Sec-2.2} and references therein) the perspective of clustering discussed in this work has not been proposed so far. Moreover, to the best of our knowledge the geodesic criterion proposed on this paper, which utilizes in a very natural and definitive manner the geometry of the underlying Wasserstein space, has not appeared in the literature before.

The discussed learning approach accompanied with the proposed geodesic clustering criteria and indices are first applied to a simulated dataset in order to illustrate the scheme's capabilities to recover the original pattern of the data. Then, two real world applications are considered, (a) clustering sovereign bonds to groups and (b) classification of pixels in satellite images, in order to provide two concrete empirical examples where the discussed framework is directly applied to uncertain and structured data and assess the performance of the geodesic clustering indices and criteria.

\section{A unified framework for clustering uncertain or structured data}\label{sec-2}

In this section we propose a unified framework for clustering uncertain or structured data based on the idea of representing such data in terms of appropriate probability measures, and using the metrization of the new feature space in terms of the Wasserstein metric. Based on the geometry of this space, a novel geodesic criterion for quantifying the homogeneity of clusters which allows for the determination of the optimal clusters number is proposed.

\subsection{Motivation and general framework}\label{sec-2.1}

Consider a collection of data points $x_1, x_2, \ldots, x_{n}$ on $\R^{d}$, corresponding to some type of measurements. The standard clustering procedure would involve the application of a clustering algorithm, e.g. the K-means algorithm, considering these observations as points on the Euclidean space $\R^{d}$, endowed with the standard Euclidean metric, and the observations will be clustered according to their distance from ficticious centers located at certain positions in $\R^{d}$, in terms of the Euclidean distance. Such clustering procedures have been proved very successful in understanding the structure of data in statistical learning procedures. Even in the case where the data can be efficiently represented as elements of $\R^d$, the choice of metric with which this space is endowed may affect the clustering result (see e.g. \cite{papayiannis2018learning}). 

However, there are many instances where the data under consideration cannot naturally be represented on a linear space, either because of special constraints or because of their collection type and sampling method. In such cases the K-means algorithm has to be modified in order to take into account the non-linearity of the underlying space. 

A large class of data types such as the ones described above, can be understood as probability measures and will henceforth be referred to as {\it measure-valued data}. In order to motivate this statement, we present below four particular examples. 

\begin{example}[{\bf Data that can be understood as probability measures}]\label{ex-1}
In many cases, we can assume that what we actually need from our observations, is not the value of the observation as such, but the collective behaviour of relevant data. For example, one may wish to consider the distribution of a physical quantity (e.g. temperature, wind speed) for a certain time period and wish to understand the differences between the distributions of the same quantity on different time periods or locations. Under this perspective, the individual data related to the particular time periods or locations will be used to obtain (empirical) estimations of the probability measures describing the quantity. Wishing to study the aggregate characteristics, which are described by the aforementioned probability measure, one may consider as observed data the induced probability measures, therefore leading to the framework of measure-valued data. Issues such as clustering or classification of these probability measures arise naturally (e.g. in remote sensing, please see Section \ref{sec-3.3}).
\end{example}

\begin{example}[{\bf Data point clouds}]\label{ex-2}
Often, spatial data can be understood as data point clouds that carry some heterogeneity. A particular example can be the onset and evolution of an epidemic, in which the spatial location and severity of several incidents is of interest. Certain spatial locations may be hit with different severity and to classify such regions into clusters and assess the spatio-temporal allocation of risk. Data point clouds can be identified as empirical probability measures using convex combination of Dirac measures of the incidents taking into account location, frequency and severity or smoothed representations using appropriate kernel-based estimators.
\end{example}

\begin{example}[{\bf Matrix-valued data}]\label{ex-3}
Other types of data can be perceived as measure-valued data, hence making feasible their treatment as elements of a Wasserstein space. An example of particular interest are matrix valued data and in particular correlation or covariance matrix data. As well known Gaussian measures on $\R^d$ are uniquely determined by their mean and their covariance matrix hence there is an one-to-one correspondence between the space of centered Gaussian measures and the manifold of positive-definite matrices (the natural space for covariance matrices) (see, e.g. \cite{masarotto2019procrustes, zemel2019frechet}). Hence, one can metrize the later using the Wasserstein metric in the space of Gaussian measures. This metric is usually called the Bures-Wasserstein metric (\cite{bhatia2019bures}) and can be calculated explicitly in terms of the typical quadratic Wasserstein distance for Gaussian measures with zero locations. As a result, the proposed algorithm in Section \ref{sec-2.2} can be directly implemented without any modifications for clustering matrix-valued data. In fact, some of the applications presented in paper can be understood as clustering matrix valued data, as for example the application on bond returns in Section \ref{sec-3.2}. 
\end{example}

\begin{example}[{\bf Uncertain data}]\label{ex-4}
In many cases the observations of a dataset are not that certain due to the effect of experimental or observational error, noise to data, inconsistencies of the monitoring equipment, etc. The uncertainty of the measurements introduces some fuzziness as to the exact location of the points on the Euclidean space. One way to take into account this fuzziness is to substitute the data points with probability measures $\mu_1,\mu_2,\ldots, \mu_n$, each one centered at its location vector $m_{i}=x_{i}\in\R^d$, with a covariance matrix $S_{i} \in \R_{+}^{d\times d}$, with ${\R}_{+}^{d\times d}$ denoting the set of symmetric and positive definite matrices of dimension $d$, modeling the uncertainty as to the exact location of the point. If geometric intuition is of any use, we are substituting each data point $x_{i} \in \R^{d}$, with an ellipsoid in $\R^{d}$, located at $m_{i}=x_{i}$, whose principal axes are indicated by the eigenvectors of the covariance matrix $S_{i}$, and we assign a probability that the actual measurement is located within this ellipsoid in terms of the probability measure $\mu_{i}$. The different dispersion matrix $S_{i}$ corresponding to the different probability measures $\mu_{i}$ model the possible differential uncertainty assigned to each measurement. This is a very reasonable and plausible assumption. For example, each measurement may be obtained by a different source or a different measurement procedure, e.g. they may correspond to measurements of a quantity performed at different geographic locations, where remoteness introduces observational errors (one can easily consider applications in observational astronomy, meteorology, medical imaging, remote sensing, etc). 
\end{example}

Motivated by the above examples we consider a collection of measures $\mu_i$ for $i=1,2,...,n$ that arose from an observation process or some statistical experiments. In particular, we will treat our observations as elements in the set of probability measures with finite moments of second order denoted by ${\cal P}_{2}(\R^{d}),$ endowed with the topology induced by the quadratic Wasserstein metric (see e.g. \cite{santambrogio2015optimal, villani2003topics}). This is a natural choice for metrizing the space of probability measures consistent with the weak-* topology, with the Wasserstein metric providing an appropriate notion of distance between two probability measures, taking efficiently into account the geometrical structure and non-linear nature of the underlying space.

We now proceed to the question of clustering these measure-valued observations. Since clustering requires a notion of distance which is appropriate for quantifying divergence between the objects under consideration, we propose the use of the Wasserstein metric space for this purpose. For example, this choice of metric space setting, can accommodate all the applications mentioned in Examples \ref{ex-1}-\ref{ex-4}. For instance, concerning Example \ref{ex-4} our proposal properly takes into account the observational uncertainty (which may be of different degrees depending on the observation) of the original measurements and when combined with the K-means clustering algorithm is expected to lead to more robust results. Note that in the limit as $S_{i} \to 0$, the probability measure $\mu_{i}$ degenerates to Dirac mass $\delta_{x_{i}}$ on $\R^{d}$, and the Wasserstein distance $W_{ 2 }(\mu_{i},\mu_{j})$  degenerates to the standard Euclidean distance $\|x_{i}-x_{j}\|_2$ in $\R^{d}$. Similar comments can be made also for the other examples.

We close the discussion by indicating why we expect that a K-means algorithm based on the Wasserstein distance leads to different results. This is best illustrated, within an example in which  in terms of a limiting procedure we may pass from a description of data in Euclidean space to its description in measure space; the example of uncertain data (Example \ref{ex-4}) is ideally suited for this purpose. Consider that at some point in the course of the clustering procedure we have two potential cluster centers $\bar{x}_{c_1}$, $\bar{x}_{c_2} \in \R^{d}$ and an observation $y \in \R^{d}$ that has to be assigned to one of these two centers. Assume without loss of generality that $\| y -\bar{x}_{c_1}\|_2 <  \| y -\bar{x}_{c_2}\|_2$ so that observation $y$ is assigned to the cluster $c_1$ centered at $\bar{x}_{c_1}$. Consider now the case where observational uncertainty is assigned to the observations. Since the potential cluster centers are obtained dynamically as centroids of subsets of the original observations, the uncertainty concerning the observations propagates to some uncertainty as to the true location of the centers $\bar{x}_{c_1}$, $\bar{x}_{c_2}$. In the framework described above, this uncertainty is modeled by considering both the observation to be assigned to a particular cluster, as well as the centers of the two potential clusters the observations belongs to, as probability measures $\mu, \bar{\mu}_{c_1}, \bar{\mu}_{c_2} \in {\cal P}_{2}(\R^{d})$, with locations and dispersions $(m,S)=(x, S)$, $(\bar{m}_{1},\bar{S}_{1})=(\bar{x}_{c_1},\bar{S}_{1})$, $(\bar{m}_{2},\bar{S}_{2})=(\bar{x}_{c_2},\bar{S}_{2})$, respectively, with the covariance matrices reflecting the extent and the nature of the observational uncertainty. 

Let us assume that these probability measures belong to the same Location-Scatter family, say without loss of generality the Gaussian family, which is a good choice for modeling observational uncertainty on account of the central limit theorem. Taking this uncertainty into account we must necessarily use the Wasserstein distance between the resulting probability measures as a criterion of assigning the point $\mu \in {\cal P}_{2}(\R^{d})$ to one of the two clusters centered at the points $\bar{\mu}_{c_1},\bar{\mu}_{c_2} \in {\cal P}_{2}(\R^{d})$. Using this criterion we will assign the point $\mu$ to the cluster $c \in \{c_1, c_2\}$  by setting $c = \arg\min_{ c \in \{c_1, c_2\} }  \{ W_{2}(\mu, \bar{\mu}_{c_1}),  W_{2}(\mu, \bar{\mu}_{c_2}) \}$. According to \cite{alvarez2016fixed}, we have that
\begin{eqnarray}\label{eqq}
W_{2}^{2}(\mu, \bar{\mu}_{c_j})=\| m- \bar{m}_{c_j}\|_2 ^2+ Tr\left( S+\bar{S}_{j}-2 (S^{1/2}\bar{S}_{j} S^{1/2})^{1/2}  \right), \,\,\, j=1,2.
\end{eqnarray}
From this expression it becomes clear how uncertainty may alter the assignment to a particular cluster. Suppose that $\| m-\bar{m}_{c_1}\|_2 <  \| m-\bar{m}_{c_2}\|_2$, so that the standard Euclidean K-means assigns this observation to cluster $c_1$. If
\begin{eqnarray*}
 Tr\left( S+\bar{S}_{1}-2 (S^{1/2}\bar{S}_{1} S^{1/2})^{1/2}  \right) > Tr\left( S+\bar{S}_{2}-2 (S^{1/2}\bar{S}_{2} S^{1/2})^{1/2}  \right),
\end{eqnarray*}
then using the Wasserstein distance the same observation will be assigned to cluster $c_2$ instead. This difference in assignment is attributed as an effect of observational uncertainty and will lead to more robust clustering of the uncertain data. As a nice geometric intuitive approach consider ellipsoids of different orientations that have to be accommodated to different clusters. Clearly, the orientation of each ellipsoid plays a role as to which cluster it is appended to. 

\subsection{Related literature and connections}\label{Sec-2.2}

In recent literature, several approaches have been discussed relying on the Wasserstein metric and Wasserstein barycenters. In \cite{irpino2014dynamic} clustering schemes are proposed for Symbolic Data Analysis, in particular for clustering histogram data, and a weight selection rule is developed for the scheme's weights relying on the one-dimensional Wasserstein metric. In \cite{del2019robust} a more sensible approach in clustering probability measures is proposed by combining results from the study of Wasserstein barycenters for the case of measures from the Location-Scatter family and the concept of trimmed means. More recently, \cite{verdinelli2019hybrid} provided a hybrid scheme for clustering probability measures where Wasserstein distance is used as a metric tool for the characteristics of a probability measure that can be modeled by Gaussian approximations while the residual terms dissimilarities are quantified by a complementary distance function on the measures projections to the tangent space to the Wasserstein space.  

Our work differentiates from the above in the following ways. In our work, clustering is examined from the perspective of {\it uncertain data}, i.e. data where the observation error could be significant and therefore the observation itself may differ from the real situation, or {\it structured data}, i.e. data for example in terms of matrices or images restricted by particular structure (positive-definite matrices). While data of this form are not measure-valued data as such, we propose a general framework for their description in terms of related probability measures, which are then treated as objects in the appropriate Wasserstein space. Second, we propose a geodesic criterion for assessing the dependence in the quality of the clustering procedure in terms of the number of clusters which can lead to a selection criterion through a scoring function. This criterion makes appropriate use of the intrinsic geometry of the metric space for the reformulated data and has not appeared in the literature before. Uncertain data or structured data of the type covered by the proposed methodology find important use in a number of interesting applications. Such circumstances occur in sciences like astronomy, remote sensing, biology, and others due to inconsistencies of the observational mechanisms. Another important situation where uncertain data appears is in financial markets where volatility significantly affect the estimation procedures for the underlying econometric models' parameters and introduces uncertainty to the estimated quantities values. 

\subsection{The Wasserstein K-Barycenters algorithm}\label{sec-2.2}

Motivated by the discussion in the previous section we consider the following problem: Given a collection of measure-valued data (which arise from complex data e.g. of the type mentioned in Examples \ref{ex-1}-\ref{ex-4}) cluster the data into like groups. On account of the nonlinear nature of the space of probability measures (or probability distributions) the standard K-means algorithm in Euclidean space is no longer sufficient for this task. On the other hand, clustering is heavily based on the concept of distance between the objects which are to be separated, hence an appropriate metrization of the space of probability measures has to be adopted before we may proceed further to our goal. A very appropriate concept of distance for quantifying divergences between different probability measures is the Wasserstein distance which is compatible with the topology of the space of probability measures (see \cite{santambrogio2015optimal, villani2003topics} for justification of this metric properties). The Wasserstein distance is in general expressed as the standard optimal transport problem originated by \cite{monge1781memoire}. Given a quadratic cost functional, for any two probability measures $\mu,\nu \in \P(\R^d)$ the Wasserstein distance is mathematically expressed as the minimization problem
\begin{equation}\label{wdist}
\mathcal{W}_2(\mu,\nu) = \inf_{T \in {\mathcal TM}(\nu) } \left\{ \int_{\R^d} \|x-T(x)\|^2 d\mu(x) \right\}^{1/2}
\end{equation}
where ${\mathcal TM}(\nu) := \{ T:\R^d \to \R^d \,\,\, :\,\,\, X\sim \mu,\,\,T(X) \sim \nu \}$ i.e. denotes the space of transport maps that {\it transform} the random variable $X\in\R^d$ which is distributed according to the probability measure $\mu$ to the random variable $Y:= T(X)\in\R^d$ distributed according to the probability measure $\nu$. In other words, the optimal transport map $T^*$ {\it pushes-forward} the measure $\mu$ to the measure $\nu$, i.e. $\nu(A) = T^*_{\#}\mu(A)$ for any Borel subset $A\subset \R^d$. 

Assume that there is a collection of $n$ probability measures $\mathcal{M} := \{\mu_{1}, \mu_2, ..., \mu_n\}$ on $\R^d$. The problem under consideration is to separate this collection of measures into $K$ clusters with the criterion of their ``affinity'' in the space of probability measures on $\R^{d}$ ($\P(\R^d)$), requiring finite second moments, ${\cal P}_{2}(\R^d)$, endowed with the metric structure induced by the quadratic Wasserstein metric, $W_{2}(\cdot, \cdot)$. As the initial step of the clustering procedure, $K$ points on $\mathcal{P}_2(\R^d)$ (probability measures) should be chosen as the initial cluster centroids (this measures could be chosen either randomly or according to some rule, e.g. the most distant measures between them). Then, the cluster membership of each measure $\mu_i \in \mathcal{M}$ is determined according to the minimal Wasserstein distance criterion of this measure from each one of the centroids (note that the minimal Wasserstein distance is equivalent to the minimal length curve connecting each point with each cluster centroid). Next, the centroids are updated according to the mean sense derived by the Wasserstein barycenter of the measures classified to each cluster, i.e. the notion of mean element in a set of probability measures employing the metric sense of the Wasserstein distance. In particular, given that the measures $\mu_{k,i}$ for $i=1,2,...,n_k$ are the measures that have been assigned to the $k$-th cluster, the new $k$-th cluster centroid is determined through the solution of the minimization problem
\begin{equation}\label{wass-bar}
\bar{\mu}_{k} = \arg\min_{\nu \in \mathcal{P}_2(\R^d)} \frac{1}{n_k}\sum_{i=1}^{n_k} \mathcal{W}_2^2(\nu, \mu_{k,i}),
\end{equation}
i.e. through the determination of the Wasserstein barycenter of the measures that have been assigned to this particular cluster (later we will refer to this probability barycenter also as the $k$-th {\it local barycenter}). In practical applications this step could be modified and instead of using $\bar{\mu}_{k}$ as the k-th cluster centroid one may choose to use as "centroid" the element of the cluster that deviates less from $\bar{\mu}_{k}$ in order to keep an element of the sample space as cluster's characterization which may provide better interpretations. General existence, uniqueness and consistency results about Wasserstein barycenters for measures on $\R^d$ have been established in \cite{agueh2011barycenters, le2017existence, kim2017wasserstein}. The above steps of the clustering algorithm are repeated until no significant change on the cluster centers or their memberships occurs. The algorithm is presented briefly below (Algorithm \ref{wk-1}).

\begin{algorithm}
\caption{{\bf Wasserstein K-Barycenters Algorithm} }\label{wk-1}
{\bf Step 0.} Set $\ell = 0$ and choose $K$ measures out of the original observations as the initial cluster centroids, $\bar{\mu}_{k}^{(\ell)}$, $k=1,\ldots, K.$ 

{\bf Step 1.} Assign each of the measures $\mu_{i}$, $i=1,\ldots, n$, to one of the $K$ clusters, choosing the cluster membership $k(i)$ according to the rule $k(i)=\arg\min_{k \in \{1,2,\ldots , K\}}  W_{2}^2\left( \mu_{i}, \bar{\mu}^{(\ell)}_{k}\right).$

{\bf Step 2.} Calculate the new cluster barycenters 
$$\hat{\mu}_{k} := \arg\min_{\nu \in \mathcal{P}_2(\R^d)} \frac{1}{n_k}\sum_{i=1}^{n_k} W_2^2\left( \nu, \mu_{k,i}^{(\ell)} \right)\,\,\, \mbox{ for }\,\,\, k=1,2,...,K,$$ 
where $\mu_{k,i}^{(\ell)}$ denotes the $i$-th measure of the $n_k$ that have been assigned to the $k$-th cluster at step $\ell$.

{\bf Step 3.} Update the new cluster centroids by choosing one of the following rules: 
\begin{itemize}
\item[{\bf 3.1}] $\bar{\mu}^{(\ell+1)}_k := \hat{\mu}_{k}$
\item[{\bf 3.2}] $\bar{\mu}^{(\ell+1)}_k := \mu_{k,*}$ where $\mu_{k,*}$ is the closest to $\hat{\mu}_{k}$ measure assigned to the k-th cluster.
\end{itemize}

{\bf Step 4.} Set $\ell := \ell+1$ and return to Step 2, until cluster centroids do not change.
\end{algorithm}

\subsection{The geometric interpretation of the clustering procedure}\label{sec-2.3}

In this subsection, we discuss the geometric interpretations that can be extracted by the clustering procedure of the Wasserstein K-Barycenters algorithm (Algorithm \ref{wk-1}). 
Following the previous discussion, each one of the $K$ cluster centroids under the proposed setting coincides with the Wasserstein barycenter of all the measures that belong to that cluster. Denote by $\bar{\mu}_k$ for $k=1,2,...,K$ the cluster centers or {\it local barycenters} and as $\bar{\mu}_0$ the barycenter of all measures or the {\it global barycenter}. Since all the measures involved belong to the same space we are able to define the {\it shortest paths} that connect these points, i.e. their {\it geodesic curves}. A very interesting characterization of these curves is provided by \cite{mccann1997convexity} called the McCann's interpolant and relies on the concept of optimal transport maps. 

According to McCann, the geodesic curve (or shortest path) that connects two measures $\mu, \nu$ in Wasserstein space can be formulated as the set of parameterized curves
\begin{equation}\label{geodesic}
\gamma(t) = \{ [(1-t)I + t T]_{\#}\mu\}_t, \,\,\,\, t\in[0,1],
\end{equation}
where $\gamma(0)=\mu$ and $\gamma(1)=\nu$, $T$ is the optimal transport map that pushes-forward the measure $\mu$ to the measure $\nu$ and $I$ denotes the identity operator, i.e. $I_{\#}\mu = \mu$. According to this construction, the minimum length connections between the global barycenter $\bar{\mu}_0$ (i.e. the Wasserstein barycenter of all probability measures under study) and the local barycenters $\bar{\mu}_k$ for $k=1,...,K$ are characterized by the parameterized curves
\begin{equation}\label{l-geodesic}
\gamma_k(t) = \{ [(1-t)I + t T^{(k)}]_{\#}\bar{\mu}_0 \}_t, \,\,\,\, t\in[0,1], \,\,\,\, k=1,...,K
\end{equation}
where $T^{(k)}$ denotes the optimal transport map that pushes forward the global barycenter to the $k$-th local barycenter, i.e. $T_{\#}^{(k)}\bar{\mu}_0 = \bar{\mu}_k$ and it holds that $\gamma_k(0) = \bar{\mu}_0$ and $\gamma_k(1) = \bar{\mu}_k$. Using these shortest path connections between the local barycenters and the global barycenter we give an attempt in characterizing the cluster membership of the probability measures involved in the clustering task which will lead to a compactness index for the shaped clusters (see Section \ref{sec-2.5}). 

Let $\mu_i$ be an arbitrary measure that needs to be assigned to one of the $K$ clusters. It is clear from the K-barycenters algorithm, that the cluster where the measure $\mu_i$ will be assigned to, is the cluster whose centroid $\bar{\mu}_k$ will have the minimal (Wasserstein) distance from $\mu_i$, or in other words, the geodesic curve that connects the measure $\mu_i$ with the specified local barycenter $\bar{\mu}_k$ must have the smallest length comparing to the lengths of each one of the other geodesic curves $\{ \gamma_k(\cdot)\}$ for all $k=1,2,...,K$. Since the measure $\mu_i$ is assigned to the $k$-th cluster, then the $k$-th geodesic curve connecting the global barycenter $\bar{\mu}_0$ with the $k$-th local barycenter $\bar{\mu}_k$ may reveal more information regarding this assignment. This extra information is revealed by the point on the geodesic $\gamma_k(t)$ that the measure $\mu_i$ is closest, i.e. the projection of measure $\mu_i$ to the geodesic curve $\gamma_k(t)$ (projection in terms of the Wasserstein distance). The closest point location alongside this particular geodesic curve is characterized by the {\it registration parameter} $\tau_i \in [0,1]$ and its value is obtained through the solution of the optimization problem ({\it registration problem})
\begin{equation}\label{reg-prob}
\tau_i := \arg \min_{t \in [0,1]} W_2^2(\mu_i, \gamma_k(t)).
\end{equation}

The registration of measure $\mu_i$ to the geodesic curve $\gamma_{k}$, can be considered as an interpolation scheme, which allows us to approximate this measure by the interpolant between the global barycenter (i.e. mean of all observations) $\bar{\mu}_0$ and the barycenter of the particular cluster to which the particular observation (measure $\mu_i$) is assigned to. Then, the registration parameter $\tau_i$ can be realized as a rough similarity index for $\mu_i$, indicating whether the derived approximation is closer to the general global barycenter $\bar{\mu}_0$ or the barycenter $\bar{\mu}_k$. In particular, $\tau_i=1$ may indicate coincidence of $\mu_i$ with the cluster centroid (or not) while $\tau_i=0$ may indicate coincidence with the global barycenter. Therefore, the value of $\tau_i$ can be conceived as a quantification of the membership of the projection of $\mu_i$ to the particular cluster $k$ it is assigned to. The membership is stronger the closer $\tau_i$ is to 1. 

\begin{figure}[ht!]
\begin{center}
\includegraphics[width = 3in]{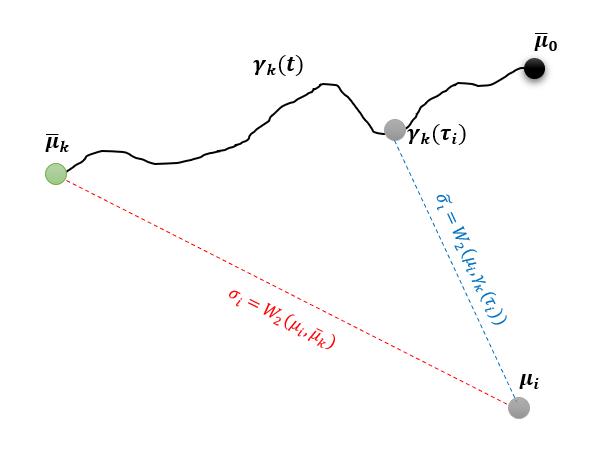} 
\caption{\it Cluster membership characterization of a point $\mu_i \in \mathcal{P}(\R^d)$ using its projections to the local geodesic curve connecting the global barycenter ($\mu_0 \in \mathcal{P}(\R^d)$) with the local barycenters that $\bar{\mu}_k$ deviates less.}\label{fig-1}
\end{center}
\end{figure}

However, the considerations above are not completely valid since the measure $\mu_i$ will not coincide with any other measure unless their distance is zero. Although the value $\tau_i$ itself may provide a first sense of similarity of the particular measure with the cluster centroid, without taking into account the distance between them, any conclusion could be proved misleading. Therefore, both $\tau_i$ and $\sigma_i := W_2(\mu_i, \bar{\mu}_k)$ (distance of the measure $\mu_i$ from the k-th local barycenter) should be taken into account in order to better characterize the membership of measure $\mu_i$. Clearly, $\sigma_i$ is the component that determines the cluster membership of $\mu_i$ through the rule $k(i):=\arg\min_{k=1,2,...,K} \sigma_i$, while the parameter $\tau_i$ provides a first sense of similarity between $\mu_i$ and $\bar{\mu}_k$. Another characteristic that contains important information is the distance between $\mu_i$ and the point's projection on the geodesic curve, i.e. $\tilde{\sigma}_i := W_2(\mu_i, \gamma_k(\tau_i))$ (see Figure \ref{fig-1}). The parameters $(\tau_i, \sigma_i, \tilde{\sigma}_i)$ contain all the important information regarding the {\it goodness of fit} of $\mu_i$ to the $k$-th cluster and its contribution to the cluster's homogeneity. On Section \ref{sec-2.5}, a cluster compactness index based on the information provided by these parameters is proposed. 

Note that $(\sigma_i, \tilde{\sigma}, \tau_i)$ endow also with a fuzzy character the whole clustering procedure. For example, assigning $\mu_i$ to a specific cluster for which the registration parameter $\tau_i$ has value close to 0, can have the interpretation that the cluster centroid is closer to this specific observation but the cluster membership is extremely ambiguous since the two measures are not that similar. On the other hand, an assignment to a cluster accompanied by a registration parameter value close to 1 means that the cluster membership looks a very natural choice. However, in the second case the value of $\sigma_i$ or $\tilde{\sigma}_i$ may chance entirely the conclusion if its value is too big (we discuss this matter later on Section \ref{sec-2.5}).

\subsection{A similarity index and clustering criteria relying on the geodesic curves}\label{sec-2.5}

In order to offer an alternative and more appropriate proposal to the existing clustering criteria, we construct an index which exploits the geodesic information revealed during the clustering procedure. From the registration problem \eqref{reg-prob} we are looking to employ the geometric information extracted by this task in order to provide a valid compactness (homogeneity) assessment of the estimated clusters by the Wasserstein K-Means algorithm discussed in \ref{sec-2.2}. In particular, consider that $K$ clusters have been derived by Algorithm \ref{wk-1} where each one contains $n_k$ probability measures where $n=\sum_{k=1}^K n_k$. For each measure which has been assigned to the $k$-th cluster (let us denote them as $\mu_{k,i}$ for $k=1,2,...,K$ and $i=1,2,...,n_k$), from the related registration procedure, there have been computed the pairs $(\tau_{k,i}, \tilde{\sigma}_{k,i})$. In order to construct a reasonable similarity index we have first to condense the information provided by these pairs to one scalar for each measure. Below we describe the approach we follow.

Assume that $n_k$ probability measures have been assigned to the $k$-th cluster with cluster centroid the measure $\bar{\mu}_k$ and the parameter triples $\{ (\tau_{i}, \sigma_{i}, \tilde{\sigma}_i) \}_i$ have been computed for each $i=1,2,...,n_k$. First, we need to characterize the element (measure) which is closest to the cluster's centroid, i.e. we define $i_* := \arg \min_{i=1,2,...,n_k} W_2(\mu_i, \bar{\mu}_k)$ and set $\mu_* = \mu_{i_*}$, $\sigma_* := \sigma_{i_*}$ and $\tilde{\sigma}_* := \tilde{\sigma}_{i_*}$. In this manner, $\mu_*$ is set as a the minimal element among the assigned measures in the $k$-th cluster in the sense that deviates less from the cluster's estimated mean tendency $\bar{\mu}_k$. Therefore, this element is reasonable to be used for the derivation of some comparison standards in order to better understand the characteristics of the other measures assigned to this cluster. The approach we follow in this attempt is through the comparison and estimation of less distant points (measures) between two geodesic curves. According to that we use the projection of each measure $\mu_i$, i.e. $\gamma_k(\tau_i)$, and we calculate its projection to the geodesic curve that connects the minimal element $\mu_*$ and the original measure $\mu_i$. Through this task we cross-validate the result we obtained by the initial registration task described in \ref{sec-2.3}.

\begin{figure}[ht!]
\begin{center}
\includegraphics[width = 3in]{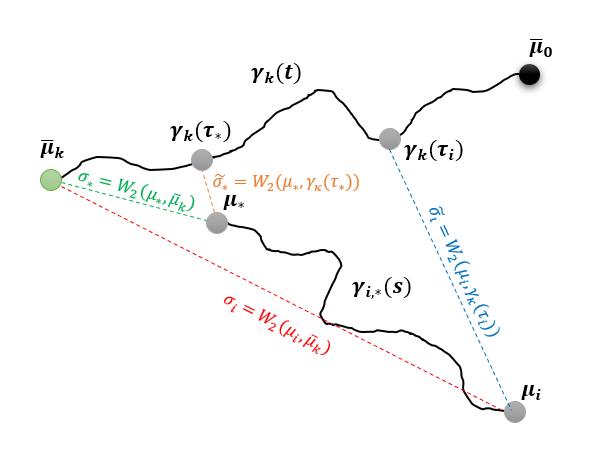} 
\caption{\it Representation of the geodesic curves involved in the double registration problem.}\label{fig-2}
\end{center}
\end{figure}

In particular, for each $\mu_i$ assigned to cluster $k$, the geodesic curve $\gamma_{i,*}(s)$ connects the measures $\mu_*$ and $\mu_i$ with $\gamma_{i,*}(0)=\mu_i$ and $\gamma_{i,*}(1)=\mu_*$. Then, the solution to the {\it reverse-registration problem}
\begin{equation}\label{reg-opt-2}
\min_{ s\in [0,1] } W_2( \gamma_{i,*}(s), \gamma_k(\tau_i) )
\end{equation}
provides the point on the geodesic curve $\gamma_{i,*}(\cdot)$ that deviates less from the point $\gamma_k(\tau_i)$ and parameterized by $s_i \in [0,1]$ which is considered as the minimizer of problem \ref{reg-opt-2} for $i$. The rationale between this second projection step is that if the initial measure $\mu_i$ is quite similar to the cluster centroid $\bar{\mu}_k$ then at the second projection we should get a point close to $\mu_*$ i.e. $s_i \to 1$. That means that the double registration procedure will reveal if the measure $\mu_i$ really belongs to the interior of the $k$-th cluster. Observe that if $\mu_i = \mu_*$ then $\gamma_{i,*}(\cdot)$ is simplified to $\mu_*$ and the above registration problem is simplified to the reverse projection of $\gamma_k(\tau_*)$ to the initial measure $\mu_*$. For $\mu_i \ne \mu_*$ the interpretation of the minimizers is of particular interest. Let us examine the extreme cases. If $s_i = 0$ and $\tau_i=0$ then measure $\mu_i$ can be characterized as an outlier (observation on the boundary of the cluster). The case where $s_i = 0$ and $\tau_i=1$ cannot happen since if the initial projection is exactly on $\bar{\mu}_k$ then the solution to the reverse registration problem has only one solution on $s_i=1$, i.e. $\mu_*$. The case $s_i = 1$ given that $\tau_i=0$ does not belong to the feasible area of the problem unless if $\bar{\mu}_k = \bar{\mu}_0$. The case $s_i = 1$ given that $\tau_i=1$ is probably the more ``difficult'' case since either the measure $\mu_i$ coincides with the minimal element, or the measure $\mu_i$ belongs to the internal area of the cluster or $\mu_i$ is an outlier which characteristics deviates more from $\bar{\mu}_0$ than $\bar{\mu}_k$. It is evident that the information obtained from the minimizing pairs $(\tau_i, s_i)$ of the double registration problem (optimization problems \ref{reg-prob} and \ref{reg-opt-2}) is not enough for the full characterization of the status of each point assigned in $k$-th cluster. Therefore we need to further enhance our distinction argument using the information provided by the distances $\sigma_i$ and $\tilde{\sigma}_i$ in order to provide a valid measure of homogeneity/compactness of the clusters. The above discussion justifies the definition of the following similarity index.

\begin{definition}[{\it Geodesic Similarity Index}]
The {\it geodesic similarity index} of a measure $\mu_i$ assigned to the $k$-th cluster with respect the cluster's centroid is defined as
\begin{equation}
g_{k,i} := s_{i}\frac{\tau_{i}}{\tau_*} \frac{\tilde{\sigma}_*}{\tilde{\sigma}_i}
\end{equation}
where $(s_{i}, \tau_{i})$ denotes the solution to the double registration problem \ref{reg-prob}-\ref{reg-opt-2} for $\mu_i$.
\end{definition}

This index is a rescaled version of the initial $\tau_i$, however taking into account also the effect of the true distance from the geodesic curve normalized with respect to the closest point to the cluster center. Clearly, for $\mu_i = \mu_*$ we obtain an index value equal to 1 since it is considered as the ideal case. In particular, due to its convenient scaling, each registration parameter $g_{k,i}$ can be realized as the level of certainty/ambiguity regarding the true membership of a particular observation ($\mu_{k,i}$) to the $k$-th cluster. Clearly, if $g_{k,i} \to 1$ there is a great amount of certainty that the $k$-th cluster is a very appropriate match for this observation, while if $g_{k,i} \to 0$ there is a great level of ambiguity that the $k$-th cluster is the most suitable choice for it. Note that, for the clusters that have been derived up to the time the registration parameters are computed, there is no doubt that the assigned observations to these clusters are rightly assigned there. However, small $g$ values by many observations indicate a cluster of low homogeneity (non-compact) which maybe reveals a poor initial decision for the number of clusters. 

\begin{definition}[{\it Geodesic Compactness Index}]
Given that $K$ different clusters have been determined and taking into account all the measures assigned to each cluster, we define the {\it Geodesic Compactness Index} (GCI) for the cluster $k$  as
\begin{equation}\label{ckindex}
GCI_k(K) := \frac{1}{n_k}\sum_{i=1}^{n_k} g_{k,i} \mbox{  for all  } k=1,2,...,K.
\end{equation} 
Moreover, a total compactness index for the total clustering procedure is naturally defined through the weighted sum
\begin{equation}\label{cindex}
GCI(K) := \sum_{k=1}^K \frac{n_k}{n} GCI_k(K) = \frac{1}{n}\sum_{k=1}^K \sum_{i=1}^{n_k}g_{k,i}. 
\end{equation}
\end{definition}

The interpretation of the above indices is in complete accordance with the mean tendency of the similarity indices. A value of the cluster index close to 1 is interpreted as an artifact of high homogeneity of the specific cluster while a an index value close to 0 is interpreted as very low homogeneity of the cluster. Observing low values to the majority of clusters determined indicates that the choice have been made for the number of clusters is maybe not appropriate. On the other hand, observing high values of the compactness indices for the majority of the determined clusters, indicates an appropriate choice for the number of clusters and success in the determination of compact clusters. However, note that if $K=n$ we get that $GCI_1(n)=GCI_2(n)=...=GCI_n(n)=1$ which is clearly not an indication of a good choice for the number clusters but rather a limit case. 

\section{Applications}

In this section, in order to illustrate the capabilities of the proposed clustering scheme and the associated geodesic criterion for the selection of the clusters' number three particular applications are considered. The first one, concerns a simulation study on the famous Fisher's iris dataset in order to provide a clustering example in artificially created uncertain data where the true cluster membership is known in order to test the scheme's behaviour in recovering the true data pattern. Next, two applications in real data are presented. The first one concerns the identification of groups of EU countries with respect to their credit profiles as represented from their governments' sovereign bond yields while the second one, concerns a standard task in remote sensing, that of image classification. In all three applications, besides the clustering scheme's resulting groups evaluation, is also illustrated the performance of the proposed compactness index.


\subsection{A study on simulated data based on Iris dataset}\label{sec-3.1}

In this section, we aim to demonstrate how the proposed clustering approach performs in retrieving the true pattern of uncertain data, considering a toy example based on the famous Fisher's iris dataset. Based on the true cluster membership of the observations, to each observation a probability distribution is assigned based on the idea we described in the introduction corresponding to each observation's uncertainty zone.  For the sake of simplicity, we work under the assumption of Gaussian measures.

\begin{figure}[ht!]
	\centering
	\begin{minipage}[t]{0.45\textwidth}
		\includegraphics[width=\textwidth]{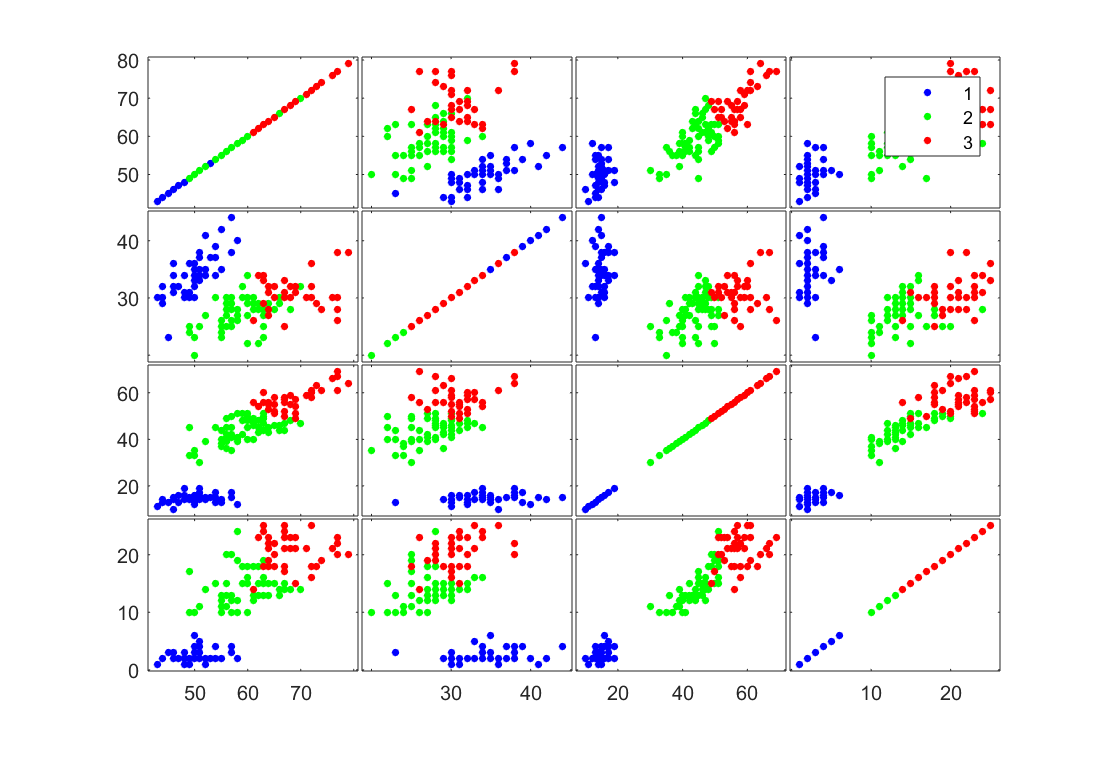}
	\end{minipage}
	\begin{minipage}[t]{0.45\textwidth}
		\includegraphics[width=\textwidth]{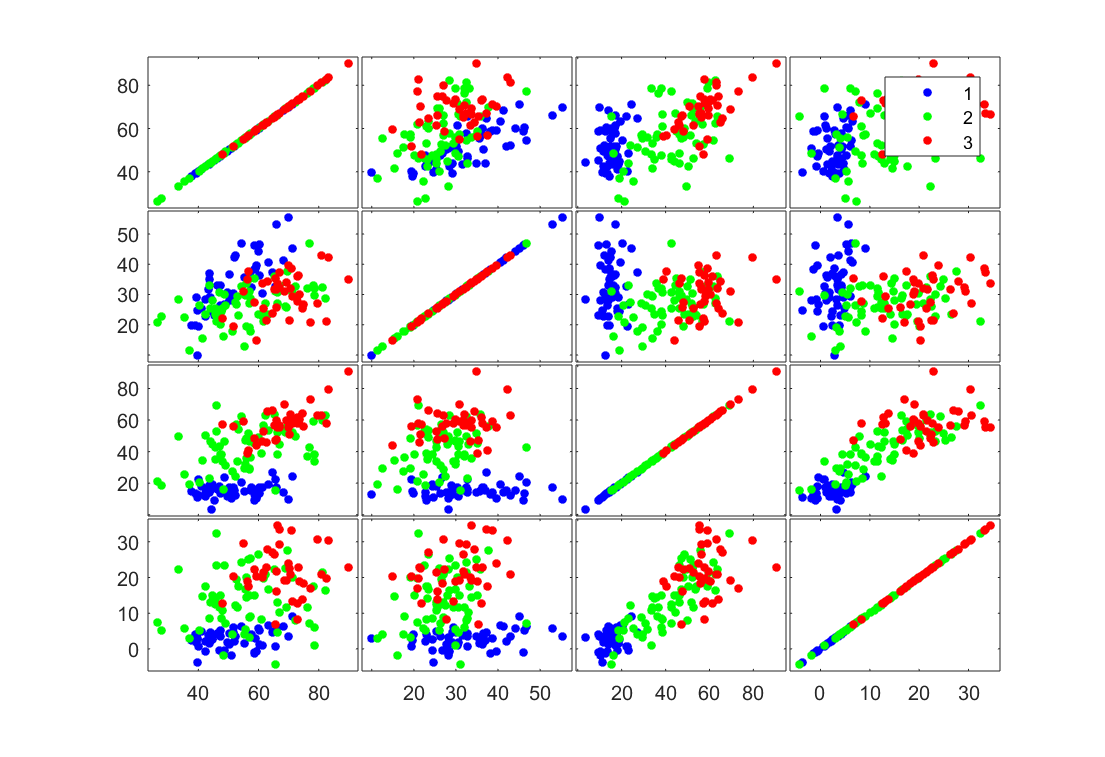}
	\end{minipage}
	\begin{minipage}[t]{0.45\textwidth}
		\includegraphics[width=\textwidth]{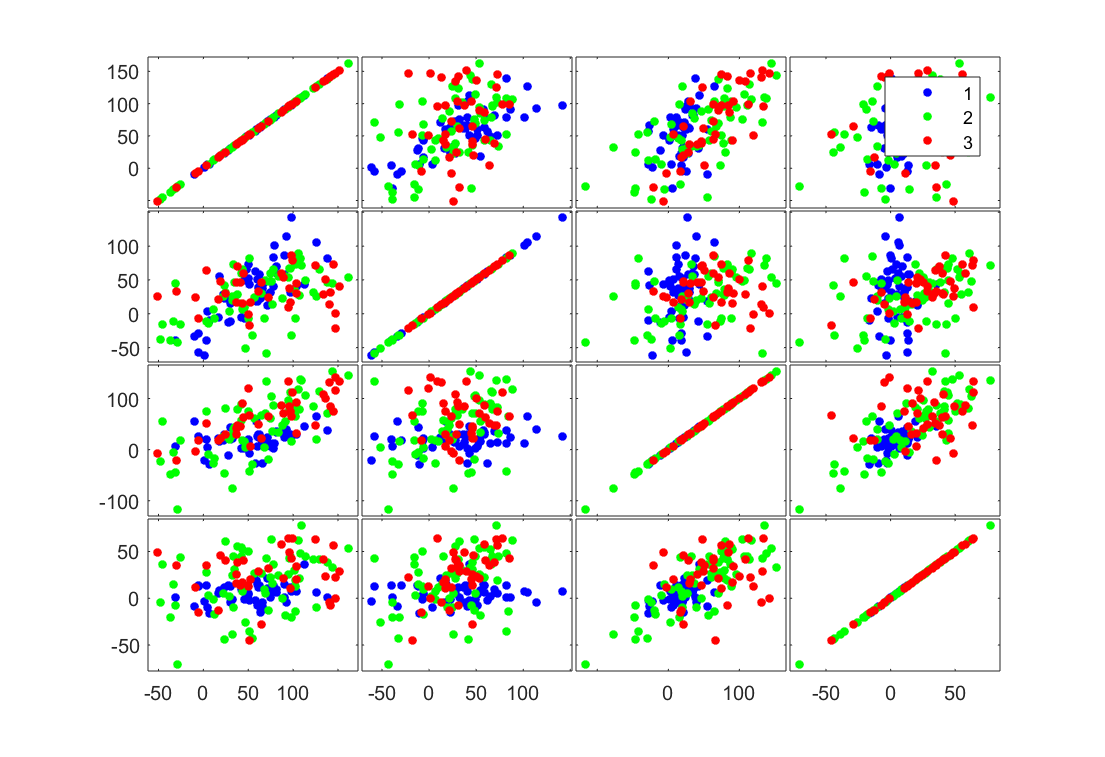}
	\end{minipage}
	\caption{Samples from iris data for increasing noise levels}\label{fig-3}
\end{figure}

In particular, let $x_i\in\R^d$ be an observation that belongs to the $k$-th cluster for $k=1,2,3$. Then, we substitute this observation with the probability measure $\mu_{x_i} = N_d(x_i, \Sigma_{x_i})$ for $i=1,...,n_k$ and $k=1,2,3$. Since, the true observation is $x_i$ it is reasonable to choose the observation itself as the location parameter. Next, a dispersion matrix $\Sigma_{x_i}$ is assigned to each observation in each cluster, however this task should be done with a carefull manner. Since the observation $x_i$ belongs to a certain cluster $k$, the distribution $\mu_{x_i}$ should have some common characteristics with the distribution of the cluster, i.e. the assigned uncertainty should be related to the uncertainty appeared in the cluster (i.e. it should be related to the cluster's dispersion matrix). Denoting by $\Sigma_k \in \mathbb{R}_{+}^{d \times d}$ the cluster's dispersion matrix, we assume that each dispersion matrix $S_{k,x_i}$ assigned to each one of the points belonging to the cluster $k$, is distributed according to the Wishart distribution with center parameter the positive definite matrix $\Sigma_k$ and degrees of freedom $n_k$, i.e. $n_k S_{k,x_i} \sim \mathcal{W}_d(\Sigma_k, n_k)$ for all $i=1,...,n_k$. 

Since the amount of uncertainty is not equal for all observations in a cluster, we further assign a scale parameter $\theta_i\in\R_+$ which is unique for each observation and expresses the distance of the location parameter $x_i$ from the cluster center $m_k$, i.e. $\theta_i = l_2(x_i, m_k)$. Then, each point $x_i$ belonging to the $k$-th cluster is replaced by the probability law $x_i \sim N_d(x_i, \Sigma_{k,x_i})$, where $\Sigma_{k,x_i} := \theta_i S_{k,x_i}$, in order to define uncertainty (fuzzyness) in a causal manner. From the probability measures we constructed above, we simulate random samples and apply the clustering approach.

\begin{table}[ht!]
	\centering
	\begin{tabular}{|l|ccccc|}
		\hline
		Sample Size & Correct Classification \% & $GCI_1(3)$ & $GCI_2(3)$ & $GCI_3(3)$ & $GCI(3)$\\
		\hline
		\multicolumn{6}{|c|}{{\it Low Noise Case} }\\
		\hline
		$B=5$   & 90.73 & 0.941 & 0.758 & 0.875 & 0.858\\%
		$B=20$ & 92.87 & 0.950 & 0.772 & 0.881 & 0.867\\%
		$B=50$ & 97.47 & 0.971 & 0.791 & 0.895 & 0.886\\%
		\hline
		\multicolumn{6}{|c|}{{\it Medium Noise Case} }\\
		\hline
		$B=5$    & 85.93 & 0.911 & 0.753 & 0.870 & 0.845\\%
		$B=20$  & 90.53 & 0.939 & 0.761 & 0.872 & 0.857\\%
		$B=50$  & 95.87 & 0.969 & 0.765 & 0.881 & 0.872\\%
		\hline
		\multicolumn{6}{|c|}{{\it High Noise Case} }\\
		\hline
		$B=5$  & 77.40 & 0.901 & 0.712 & 0.820 & 0.811\\%
		$B=20$ & 84.27 & 0.922 & 0.754 & 0.843 & 0.840\\%
		$B=50$ & 93.40 & 0.957 & 0.764 & 0.873 & 0.865\\%
		\hline
	\end{tabular}
	\caption{Clustering diagnostics for the noisy scenarios considered}\label{tab-1}
\end{table}

Obviously, in this toy example, the typical pointwise clustering approaches will fail since, from one sample (drawing one sample from the probability measures constructed above) it cannot reveal any information regarding the characteristics of each measure, and therefore the pointwise approach fails. More than one samples from the dataset needs to be collected in order to at least obtain some estimates (even rough estimates) of the special characteristics of each distribution such as the location and dispersion. At this point, we introduce a scaling parameter $\kappa \in [0,1]$ which controls the noise effect in data, by assuming now that each sample is drawn from $x_i \sim N_d(x_i, \kappa \Sigma_{k,x_i})$. For $\kappa = 0$ we retrieve the original dataset while for the choices $\kappa = 0.1, 0.5, 1$ we define the ``low'', ``medium'' and ``high'' noise cases, respectively. In order to assess the ability of the proposed method to recover the true cluster memberships of the noisy data we simulate samples of size $B= 5, 20$ and $50$ in all noise levels. The results are illustrated in Table \ref{tab-1} where the ability of the proposed clustering method to recover the true cluster membership from noisy data at a very notable percentage is displayed. The Wasserstein K-Barycenters approach demonstrates significant learning ability in retrieving the true clusters in all cases since it takes into account the geometrical characteristics of the data leading to accelerated accuracy of the clustering algorithm as data sample grows ($B\to \infty$). It is also remarkable the ability of the algorithm to construct very compact clusters since all compactness indices has values greater that 0.7, indicating the existence of very compact clusters.


\subsection{Clustering EU Bond Yields}\label{sec-3.2}

As a first illustrative application in real data we consider the task of clustering the countries of eurozone according to their credit profiles for the time period 2001-2019. The data under consideration is the weekly yield rates for each country's zero coupon government bonds with maturity horizons from one to ten years. The data are collected, on a monthly basis, through the official DataStream database and the countries for which the clustering task is performed are those which were members of the European monetary union constantly for the time period 2001 to 2019. Note that data for Luxemburg are not available (possibly not such type of bonds have been published from its government) therefore it is excluded from the analysis. As a result, the countries included in the analysis (followed by their abbreviations) are: Austria (AT), Belgium (BL), Finland (FI), France (FR), Germany (DE), Greece (GR), Ireland (IE), Italy (IT), Netherlands (NL), Portugal (PT) and Spain (ES). 

\subsubsection{Credit Profile Determination through Probability Measures}\label{sec-3.2.1}

The available data used for the clustering task are for each particular country included in the study, the observed bond indices for each different maturity horizon as it is recorded at the end of each month. For example, for a two year period, for the country $i$ there are available $104$ records for the published by the country's government bonds with maturity horizon $m=1,2,...,10$ years. Therefore, a record at the specified time $t$ (i.e. at the end of a certain week in the time period under consideration) is the vector
$$ {\bm r}_t^{(i)} = \left(r^{(i)}_{t,1}, r^{(i)}_{t,2}, ..., r^{(i)}_{t,10}\right)'\in \R^{10} $$
where each $r^{(i)}_{t,m}$ for $m=1,2,...,10$ denotes the recorded index for the bond with maturity horizon $m$ years. As a result, if the period under consideration consists of $n$ weeks, then for each country there is available a $(n \times 10)$ data matrix with the available records. Note that although these data could be also considered as time series data, we do not employ this approach here since it is irrelevant and outside the the scope of the present work. However, we indicatively refer to the approaches in the literature that consider the matter of clustering time series data \cite{vlachos2003wavelet, niennattrakul2007clustering, huang2016time}. In particular, in this context we realise these records as realizations of a 10-dimensional random value which can be sufficiently represented by a Location-Scatter probability measure with a certain location and dispersion matrix.

Given that the economies are in a steady state, no major variations in these records should be observed in each month. For this reason, the entire time period 2001-19, that data are available, is split into six smaller periods: 2001-04, 2005-08, 2009-11, 2012-14, 2015-17 and 2018-19 in order to not allow very lengthy time periods where important changes may occur and significantly affect each country's economy. These periods represent a growth period (2001-04), followed by two financial crisis intervals: 2005-08, which is the period up to the Lehman brothers collapse, and 2009-11 that is the immediate aftermath, including the ensuing debt crisis in Greece, Portugal and Ireland. These are followed by a period when certain eurozone countries were under Economic Adjustment programmes (bailout programmes, 2012-14), a period following these programmes (2015-17) and the last two recovery years (2018-19). It is evident that one may try different time intervals (6 months, 1 year periods, etc.) or use more frequent data (e.g. daily, hourly, etc.) however since this is not the main purpose of this paper we attempt to keep such details to the most convenient for the analysis and the reader framework.

Clearly, the provided data in the current form is quite hard to analyze. Therefore, we attempt for each one of the defined time periods, to condense the available data for each country in order to sufficiently and more conveniently represent the provided information. That task is performed through the identification of each data matrix by a probability measure of the Location-Scatter family of distributions, for convenience we restrict ourselves to the family of Gaussian measures. Under this approach, for a certain data matrix (i.e. provided data for a specific country $i$) the location vector ${\bm m}_i\in\R^{10}$ represents the mean tendency of the bond indices for the particular period of time therefore it is reasonable to set 
$${\bm m}_i={\bar{\bm r}}_i=(\bar{r}_{i,1}, \bar{r}_{i,2},..., \bar{r}_{i,10})' \in \R^{10}$$ 
where $\bar{r}_{i,m} = \mathbb{E}[r^{(i)}_{t,m}] \in \R$ for $m=1,2,...,10$ denoting the mean tendency of the bond index of maturity horizon $m$ for the country $i$. Another important characteristic of the dataset is the correlation structure between the bonds of different maturity and of course their dispersion. Under the assumption that the nature of the data does not change significantly during our sample period, then the dispersion characteristics can be sufficiently estimated by the covariance matrix
$$ S_i = \mathbb{E}[({\bm r}^{(i)}_{t}-\bar{\bm r}_{i})({\bm r}^{(i)}_{t}-\bar{\bm r}_{i})']\in\R^{10 \times 10}_{+}. $$
Under this setting, the country's $i$ profile can be identified with the Gaussian probability measure $\mu_i=N({\bm m}_i, S_i)$ by condensing the information provided by the data set to location and dispersion-correlation characteristics. This is a very natural approach in most applications considering multivariate data however, one may consider also some family of probability distributions capturing asymmetries in data (e.g. through a Copula, a meta-distribution, a vine-Copula construction, etc.). However, since in this work we investigate the case of the Location-Scatter family of distributions we feel that such a consideration, although extremely interesting, it is beyond the scope of this paper. Therefore, under the above considerations, the credit profile of a country induced by the observed bond yield data can be represented by a Gaussian measure where the location and dispersion parameters efficiently condense the available information provided for the time period under consideration and the nature of the Gaussian law reflects the assumption regarding consistency, i.e. the time period length is chosen so that no dramatic changes happen in the economies of the countries under study for its duration. 

\begin{figure}[ht!]
\centering
\includegraphics[width=6in]{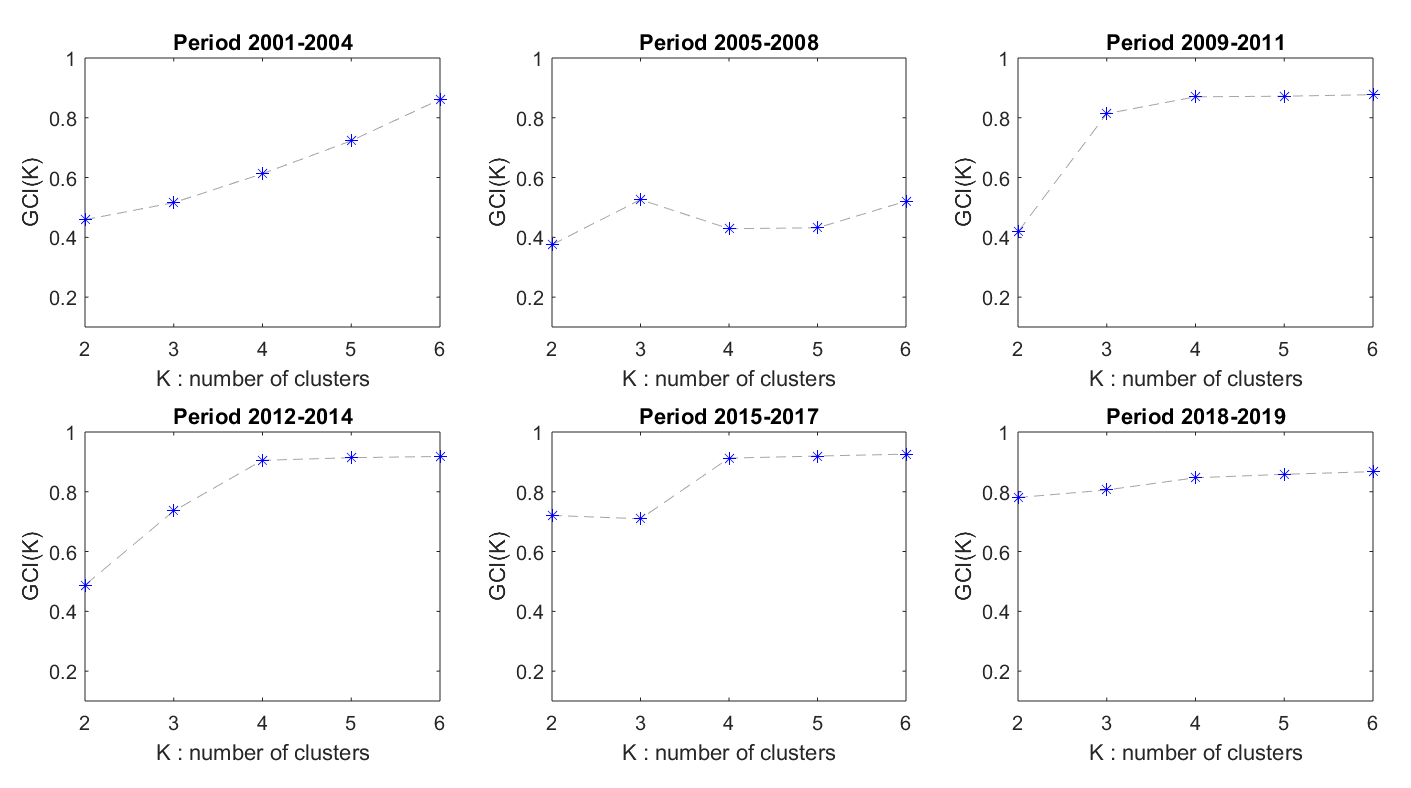}
\caption{Geodesic Compactness Index values per time period}\label{fig-3.1}
\end{figure}

\subsubsection{Clustering bond yield data with Wasserstein barycenters}\label{sec-3.2.2}

Following the aforementioned framework and the learning methodology described in Section \ref{sec-2.2}, we perform the clustering analysis of the credit profiles of the eurozone countries. The clustering task is performed repeatedly for the defined six time periods 2001-04, 2004-08, 2009-11, 2012-14, 2015-17 and 2018-19. First, we employ the geodesic criterion presented in Section \ref{sec-2.5} in order to determine a reasonable number of clusters for each time period. The results of the criterion output are graphically represented in Figure \ref{fig-3.1}. We consider up to six different clusters for each time period, since more clusters may not have any interpretative value. The rationale beyond the selection of an appropriate number of clusters in each case is to choose the number of groups where a high value (comparing to the values for other group numbers) in the compactness is succeeded and adding more groups this number is not significantly increased. Having this in mind, the optimal cluster numbers per time period are: 4 for 2001-04, 3 for 2005-08, 4 for 2009-11, 4 for 2012-14, 4 for 2015-17 and 3 for 2018-19.

\begin{table}[ht!]\tiny
\centering
\begin{tabular}{|c|c|c|c|c|c|c|}
\hline
&&&&&&\\ 
{\bf Time Period} & {\bf 2001-04} & {\bf 2005-08} & {\bf 2009-11} & {\bf 2012-14} & {\bf 2015-17} & {\bf 2018-19}\\
\hline 
&&&&&&\\
{\bf Group 1} & DE             &  AT BL DE FI & AT BL DE FI    & AT BL DE FI   & AT BL DE FI  &   AT BL DE FI\\
                    &                    &  FR NL          & FR NL             &  FR NL          & FR IE NL       &   FR IE NL\\
&&&&&&\\
{\bf Group 2} & AT              &  GR IT         & ES IT              & ES IE IT        &  ES IT          & ES PT\\
                    &                    &       &           &                     &                   & \\ 
&&&&&&\\                  
{\bf Group 3} &BL ES FI FR  &  IE              & IE PT              & PT             & PT             & GR IT\\
                     & NL &&&&&\\
 &&&&&&\\                
{\bf Group 4} & GR IE IT PT  &                  & GR                  & GR             & GR             & \\
&&&&&&\\ 
\hline
\end{tabular}
\caption{Optimal clusters according to GCI interpretation}\label{tab-3.1}
\end{table}

It seems that for the period 2001-2008 for most choices of clusters' number, the compactness index is not higher than 0.65 (higher index value are obtained for the period 2001-04 for $K>4$ however this happens because creates clusters of a single country) indicating not very high homogeneity in the clustering output. On the other hand, the clustering outputs in the period 2009-2019 is characterized by quite high homogeneity with the majority of cases (for $K>2$) succeeding a compactness index close to or above 0.8. For each time period, the obtained clusters are illustrated in Table \ref{tab-3.1} and the compactness index of each cluster in Table \ref{tab-3.2}.

\begin{table}[ht!]\small
\centering
\begin{tabular}{|l|rrrrrr|}
\hline
                    & {\bf 2001-04}  & {\bf 2005-08} & {\bf 2009-11} & {\bf 2012-14} & {\bf 2015-17} & {\bf 2018-19}\\
\hline                    
{\bf Group 1} & 1.00 (1) & 0.41 (6)  & 0.94 (6) & 0.95 (6) & 0.92 (7)& 0.90 (7)\\ 
{\bf Group 2} & 1.00 (1)& 0.75 (2) & 0.82 (2)& 0.76 (3)& 0.79 (2)& 0.51 (2)\\
{\bf Group 3} & 0.58 (5) & 1.00 (1) & 0.64 (2)& 1.00 (1)& 1.00 (1)& 0.78 (2)\\
{\bf Group 4} & 0.46 (4) &     - (0)  & 1.00 (1)& 1.00 (1)& 1.00 (1)&     - (0)\\
\hline
{\bf Total }    & 0.61 (11)  & 0.53 (11) & 0.81 (11) & 0.90 (11) & 0.91 (11) & 0.81 (11)\\
\hline
\end{tabular}
\caption{GCI values per time period and cluster}\label{tab-3.2}
\end{table}

\begin{table}[ht!]\small
\centering
\begin{tabular}{|c|ccccccccccc|}
\hline
& AT & BL & DE & ES & FI & FR & GR & IE & IT & NL & PT\\
\hline
2001 & AAA & AA+ & AAA & AA+ & AA+ & AAA & A & AAA & AA & AAA & AA\\
2002 & AAA & AA+ & AAA & AA+ & AAA & AAA & A & AAA & AA & AAA & AA\\
2003 & AAA & AA+ & AAA & AA+ & AAA & AAA & A+ & AAA & AA & AAA & AA\\
2004 & AAA & AA+ & AAA & AAA & AAA & AAA & A & AAA & AA- & AAA & AA\\
2005 & AAA & AA+ & AAA & AAA & AAA & AAA & A & AAA & AA- & AAA & AA-\\
2006 & AAA & AA+ & AAA & AAA & AAA & AAA & A & AAA & A+ & AAA & AA-\\
2007 & AAA & AA+ & AAA & AAA & AAA & AAA & A & AAA & A+ & AAA & AA-\\
2008 & AAA & AA+ & AAA & AAA & AAA & AAA & A & AAA & A+ & AAA & AA-\\
2009 & AAA & AA+ & AAA & AA+ & AAA & AAA & BBB+ & AA & A+ & AAA & A+\\
2010 & AAA & AA+ & AAA & AA & AAA & AAA & BB+ & A & A+ & AAA & A-\\
2011 & AAA & AA & AAA & AA- & AAA & AAA & CC & BBB+ & A & AAA & BBB-\\
2012 & AA+ & AA & AAA & BBB- & AAA & AA+ & B- & BBB+ & BBB+ & AAA & BB\\
2013 & AA+ & AA & AAA & BBB- & AAA & AA & B-  & BBB+ & BBB  & AA+ & BB\\
2014 & AA+ & AA & AAA & BBB  & AA+ & AA & B   & A    & BBB- & AA+ & BB\\
2015 & AA+ & AA & AAA & BBB+ & AA+ & AA & CCC+& A+   & BBB- & AAA & BB+\\
2016 & AA+ & AA & AAA & BBB+ & AA+ & AA & B-  & A+   & BBB  & AAA & BB+\\
2017 & AA+ & AA & AAA & BBB+ & AA+ & AA & B-  & A+   & BBB  & AAA& BBB-\\
2018 & AA+ & AA & AAA & A-   & AA+ & AA & B+  & A+   & BBB  & AAA& BBB-\\
2019 & AA+ & AA & AAA & A    & AA+ & AA & BB- & A+   & BBB  & AAA& BBB\\
\hline
\end{tabular}
\caption{S\&P yearly credit ratings for the EU countries for the time period 2001-2019}\label{fig-3.2}
\end{table}

From early on (2001-04), our proposed framework and methodology appears able to identify a cluster of the so--called PIIG countries (Portugal, Ireland, Italy and Greece), whose bond yields indicated back then that a debt crisis might ensue. During the eurozone debt crisis (2009-11) and the resulting bailout programmes (2012-14) Greece, the country with the most acute debt problem, is singled out as a different cluster, while Ireland and Portugal (countries under bailout programs) are grouped into a separate cluster. After the end of the bailout programs in all countries (2018-19), Ireland has ``moved'' to the core eurozone Group 1, Greece and Italy remain the countries with the highest debt--to--GDP in the eurozone (Group 3), while the Iberia countries (Spain and Portugal) constitute a separate cluster in--between.
In order to assess the validity of the obtained clustering outputs, in Figure \ref{fig-3.2} are illustrated the credit rankings of each eurozone country for the period 2001-19 on a yearly base as determined by Standard and Poors (S\&P). It is evident, that the clustering approach we followed succeeded in retrieving the true situation in most of the cases if we consider the rating mechanism of S\&P as an expert opinion.  


\subsection{Image classification with Wasserstein barycenters in remote sensing}\label{sec-3.3}

\begin{figure}[ht!]
	\centering
	\includegraphics[width=3.3in]{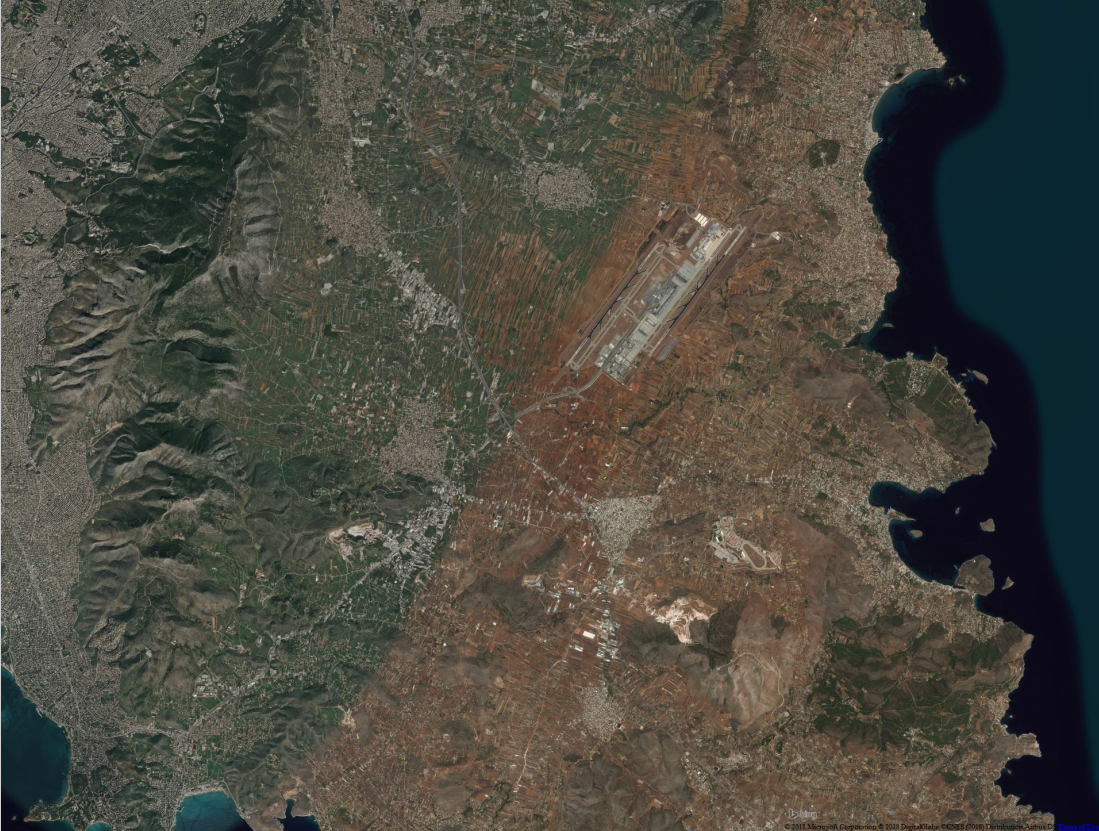}
	\caption{The satellite image from the Mesogeia area}\label{fig-3.3}
\end{figure}

In this second application, we employ the learning approach described in Section \ref{sec-2} in a standard problem in remote sensing, that of characterizing satellite images, i.e. assigning the pixels of satellite images to certain category types, e.g. built environment, agricultural environment, etc. The general practice in this task is to use as trainset for each category some sampled pixels from previous classification tasks in order to train the classification mechanism, and then use the trained system to characterize the areas of the new image. In the following, we first describe the proposed framework and then the discussed classification approach is implemented for the full characterization of a satellite image taken from the area of Mesogeia, Attica, Greece (Figure \ref{fig-3.3}).

\subsubsection{Description of the framework}\label{sec-3.3.1}

Assume that an image is available in resolution $r \times c$, i.e. it consists of $M := r\times c$ pixels. Then each pixel $p_i$ for $i=1,2,...,M$ is identified by a specific pair of coordinates $(x_i,y_i)$ that describe the pixel's position in the image and by a specific vector $p_i \in \R^d$ representing the pixel's brightness on each color band according the color band system is used, e.g. if the RGB color band system is used, then $d=3$ and $p_i \in \R^3$ for all $i$. At the time when the satellite image is taken, each pixel belongs to a specific environment type (e.g. built environment, agricultural, green areas, etc), however this category is not a priori known but it must be inferred from the available information, i.e. the values in the different color bands of the pixel and from what lies in its neighbourhood. In order to determine the environment type of each pixel, statistical methods are employed relying on databanks with classified pixels in order to characterize the new image. Below, following the discussion in Section \ref{sec-2} regarding the learning capabilities of Wasserstein barycenters, a classification approach is presented relying on a distributional-wise framework of the data unlike other statistical methods relying on local criteria like least-squares or the maximum likelihood principle . 

Assume that a train set of $M$ pixels of $K$ different environment types is available in order to properly prepare a classification scheme for the identification of the unknown areas of a satellite image. Denote by $\{p^k_i\}_{i=1}^{M_k}$ the classified sample belonging to the $k$-th environment type, where $M= \sum_{k=1}^K M_k$. Then, for each pixel $p^k_i$, $i=1,2,...,M_k$ of any category $k=1,2,...,K$, we define a neighborhood around it (for a given radius) and define the parameters 
$$ m^k_i := p^k_i \in \R^d,$$
describing the {\it location} of the pixel in the color band system that is used, and
$$ S^k_i := \frac{1}{M^{(i)}_k}\sum_{j=1}^{M_k^{(i)}}(p^k_j-m^k_i)(p^k_j-m^k_i)^T \in \R^{d \times d}_{+},$$
describing the {\it dispersion} in color band values in the pixel's neighborhood where $M_k^{(i)}$ is the number of the pixels in the neighborhood of $p^k_i$. Naturally, each pixel $p^k_i$ can be identified by a probability measure of the Location-Scatter family, $\mu^k_i = LS(m^k_i, S_i^k)$ (e.g. a Gaussian probability measure on $\R^d$). Realizing each pixel as a probability measure with the above manner, besides the information regarding the location of the specific point, are also parameterized the characteristics of its neighbourhood around it through the dispersion matrix $S_i^k$. Quantifying the information in the neighbourhood of the pixel through the covariance matrix treats also the problem of different orientation. Consider for example two pixels $p_A$ and $p_B$ where the first one has on its north neighbourhood sea and urban structure anywhere else, while the second one has sea on its east neighbourhood and urban structure anywhere else (i.e. same neighbourhood but rotated). Under the Wasserstein distance and the covariance matrix both pixels are classified in the same category regardless the different orientation of its neighbourhood (no permutations needed). 

In the case of Gaussian probability laws, the pixels of the $k$-th environment type, are identified by an appropriately assigned collection of Gaussian probability measures $\mathcal{M}_k = \{ \mu^k_i\}_{i=1}^{M_k}$. Since, all these probability measures parameterize location and dispersion characteristics of pixels belonging to the same category, we need to characterize this category's most typical element. Under the probabilistic framework we work, we employ the concept of Wasserstein barycenter in order to define the mean tendency of a pixel that should be assigned to this particular category. Then, for each environment type $k$, the corresponding probability barycenters are estimated in order to obtain a prototype for each environment, i.e. for each category $k$, the barycentric probability measure is
$$ \bar{\mu}_k = \arg\min_{\mu \in \mathcal{P}} \frac{1}{M_k}\sum_{i=1}^{M_k} \mathcal{W}_2^2(\mu, \mu_i^k)$$
where $\bar{\mu}_k \in \mathcal{P}(\R^d)$ remains a probability measure of the Location-Scatter family where its parameters can be calculated semi-analytically.

Following the above procedure, the training dataset (i.e. the classified samples) provides $K$ characteristic barycenters and this summarized information can be used in order to characterize any unclassified image according to these $K$ specified environment types. In this case, the above procedure is repeated in order to identify each pixel $p_j$ in the uncharacterized image by a probability measure $\mu_j \in LS$ (working as above with the pixels in its neighbourhood). Then, these pixels are classified to a specific environment type according to the rule 
$$k(j) := \arg\min_{k\in\{1,2,...,K\}} \mathcal{W}_2^2(\mu_j, \bar{\mu}_k). $$
Therefore, each unclassified pixel $p_j$ is classified to the environment type $k$ that the specific pixel's identified measure deviates less from the category's corresponding barycenter.

\subsubsection{A case study}\label{sec-3.3.2}

As a practical illustration of the described methodology we characterize a satellite image taken from the area of Mesogeia, Attica, Greece (Figure \ref{fig-3.3}). The colour band system used is a 6-band system, i.e. $p_i \in \R^6$. For simplicity, only four environment types are used and specifically: 1 - built environment (red), 2 - agriculture (yellow), 3 - green areas (green) and 4 - sea (blue). As a training dataset are used four samples, one for each environment type of corresponding sizes $n_1=294$, $n_2=1214$, $n_3=611$ and $n_4 = 105$. Using these samples as a training base we characterize the unclassified image using the approach described in Section \ref{sec-3.2.1}. The classification accuracy is confirmed through $N=3031$ pixels of known classification ($176$ of urban type, $1842$ of agricultural type, $159$ from sea and $854$ of green areas). For this first task, different neighbourhood sizes are used for the derivation of location and dispersion parameters for each pixel. In particular neighbourhood sizes of 8 pixels (8-p), 24 pixels (24-p), 48 pixels (48-p), 96 pixels (96-p), 192 pixels (192-p) and 4096 pixels (4096-p) are considered. The classification results (please see Table \ref{tab-3.3}) indicate that the chosen neighbourhood size do not affect that much the classification task, at least in the case where each environment type is represented through a single barycenter. In general, with that choice (single barycenters) a total accuracy around $83\%$ is succeeded for all neighbourhood sizes under consideration. However, it is evident that is some categories like Agriculture and Green Areas the individual accuracy is not very high which could be the effect of low representative power of the specific categories through the use of single barycenters.

\begin{table}[ht!]
	\centering
	\begin{tabular}{|c|cccc|c|}
		\hline
		&\multicolumn{4}{|c|}{ Accuracy per Environment Type } & \\
		Neighbourhood size & Built Env.& Agriculture & Green Areas & Sea & Total Accuracy\\
		\hline
		8-p         & 81.25\% & 77.56\% & 71.43\% & 100.00\%  & 82.56\%\\%
		24-p       & 82.39\% & 78.64\% & 71.25\% & 100.00\%  & 83.07\% \\%
		48-p       & 81.82\% & 78.50\% & 71.43\% & 100.00\%  & 82.94\%\\%
		96-p       & 81.82\% & 78.53\% & 71.10\% & 100.00\%  & 82.86\%\\%
		192-p     & 82.29\% & 78.72\% & 70.96\% & 100.00\%  & 82.99\%\\%
		4096-p   & 82.56\% & 78.98\% & 69.77\% & 100.00\%  & 82.83\%\\%
		\hline
	\end{tabular}
	\caption{Classification results using single barycenters for different neighbourhood sizes}\label{tab-3.3}
\end{table}

As a next step in our analysis we employ more barycenters for each category in order to increase the representability of the category's variability through more factors (local barycenters). For example, it seems that the information of the agricultural environment is not quantified efficiently through the use of a single barycenter. For this reason, the use of more barycenters for the representation/characterization of the same category could enhance the classification performance of the method since different fragments of information are expected to be captured by different local barycenters. Therefore, before the classification task, a local clustering step for each one of the environment types should reveal better options for the representation of each category (through more barycenters). This is a quite rational approach since the different categories present different homogeneity, e.g. sea environment is more homogeneous since a pixel of this category in its neighbourhood could possibly have either sea or coast (i.e. built environment) therefore maybe two clusters could properly model this distinction. On the other hand, environment types like agricultural should need more than one clusters (and consequently more barycenters) to properly describe the provided information since present high levels of heterogeneity due to the variety of their possible neighbours, e.g. a farm pixel could be neighbour with urban, forest or sea environment or with any combination of them. As a result, the use of more clusters (barycenters) for the description of more heterogenous environment types seems a very justified approach.

\begin{figure}[ht!]
	\centering
	\includegraphics[width=6.2in]{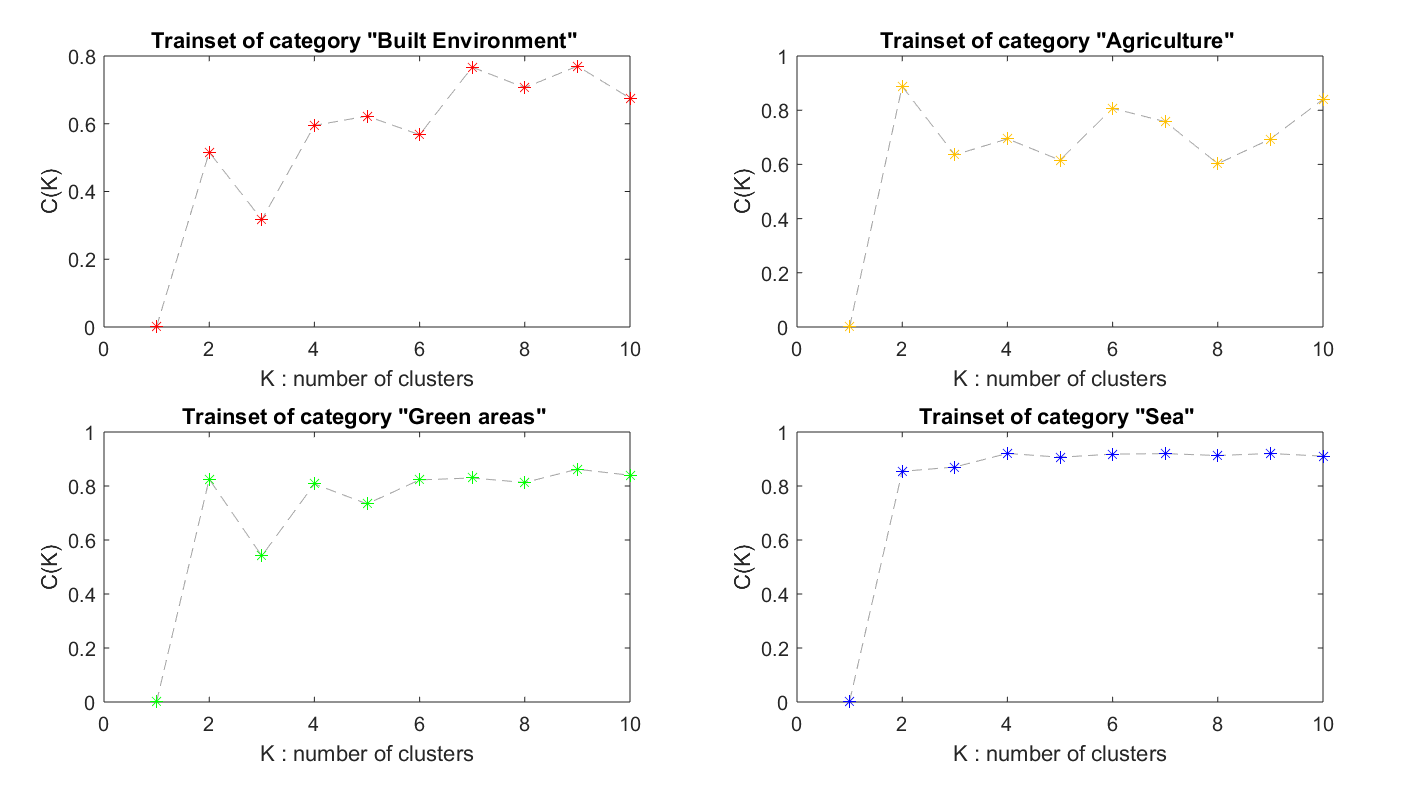}
	\caption{Geodesic Compactness Index values per environment type}\label{fig-3.4}
\end{figure}

However, there is a question on how to optimally select an appropriate number of clusters (i.e. barycenters) for each environment type. Although, the derivation of a universal optimal criterion is not possible since different types of data should need different treatment, the compactness criterion discussed in Section \ref{sec-2.5} could offer a very convenient and plausible mean of discrimination. This compactness index can be used to provide an intuition for the minimum number of clusters that are needed in order to model the information of each environment type by creating homogeneous (compact) clusters. In Figure \ref{fig-3.4} are illustrated plots for the value of index $GCI(K)$ for different choices of clusters (for $K_j=1,2,...,10$) for each environment type $j=1,2,3,4$. It seems that for the built environment  type at least two barycenters should be used while the same number of barycenters looks sufficient also for the sea category. Also, for the other two categories (agricultural environment and green areas) it seems that the minimum number of barycenters that should be used is at least two, however the use of more than two barycenters is also a very natural choice. Therefore, the occurring classification scheme should represent each environment type with more than one centroids (barycenters) which although different lead to the same classification result if an uncharacterized area is closer to just one of these centroids with respect to centroids from other environment types. For example, if on the training task, each environment type $j$ (for $j=1,2,3,4$) is divided into $K_j > 1$ sub-categories, then the final classification scheme should consist of $K = \sum_{j=1}^4 K_j > 4$ categories, however the final category assignment concerns only the initial $4$ environment types.

\begin{table}[ht!]
	\centering
	\begin{tabular}{|l|rrrr|r|}
		\hline
		&\multicolumn{4}{|c|}{ Accuracy per Environment Type } & \\
		Barycenters Model & Built Env.& Agriculture & Green Areas & Sea & Total Accuracy\\
		\hline
		$K_{1111}=4 $   & 81.82\% & 78.50\% & 71.43\% & 100.00\% &  82.94\%\\%
		$K_{2222}=8 $   & 83.52\% & 80.35\% & 87.59\% & 100.00\% &  87.87\%\\%
		$K_{3333}=12$  & 91.48\% & 79.97\% & 77.87\% & 100.00\% &  87.33\%\\%
		$K_{4444}=16$  & 90.91\% & 82.79\% & 89.70\% & 100.00\% &  90.85\%\\%
		$K_{5555}=20$  & 89.77\% & 78.07\% & 87.47\% & 100.00\% &  88.83\%\\%
		$K_{6666}=24$  & 90.34\% & 86.32\% & 85.83\% & 100.00\% &  90.62\%\\%
		$K_{7777}=28$  & 88.64\% & 86.05\% & 87.35\% & 100.00\% &  90.51\%\\%
		$K_{8888}=32$  & 92.05\% & 83.98\% & 92.39\% & 100.00\% &  92.11\%\\%
		\hline
		$K_{3582}=17$ & 86.93\% & 82.25\% & 94.61\% & 100.00\% & 90.95\%\\%
		$K_{4781}=20$ & 88.64\% & 85.07\% & 93.09\% & 99.37\%  & 91.54\%\\%
		$K_{4782}=21$ & 88.64\% & 85.07\% & 92.97\% & 100.00\% & 91.67\%\\%
		\hline
	\end{tabular}
	\caption{Classification results using multiple barycenters models for 48-p neighbourhood size}\label{tab-3.4}
\end{table}

\begin{figure}[ht!]
	\centering
	\begin{minipage}[t]{0.32\textwidth}
		\includegraphics[width=\textwidth]{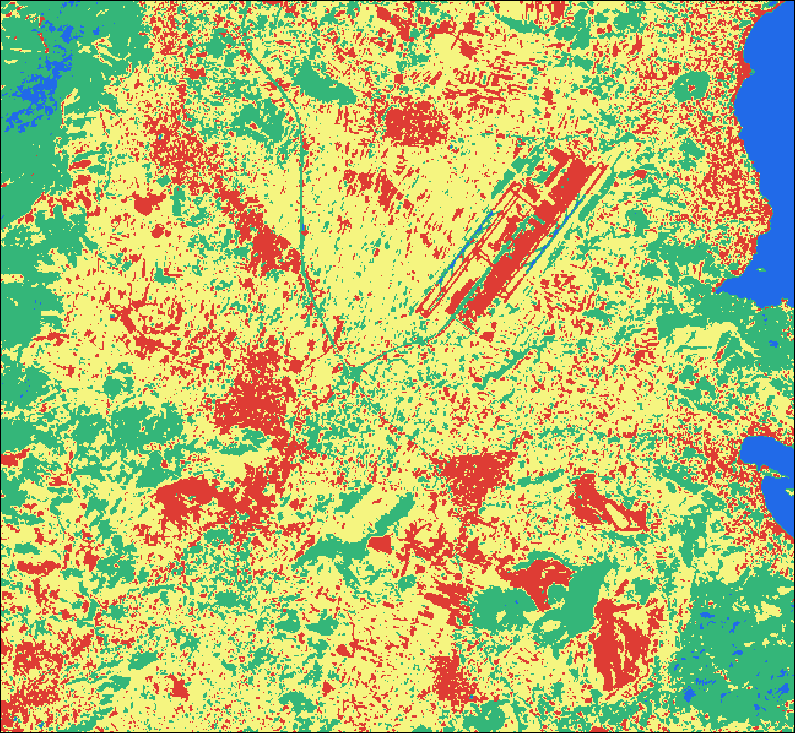}
	\end{minipage}
	\begin{minipage}[t]{0.32\textwidth}
		\includegraphics[width=\textwidth]{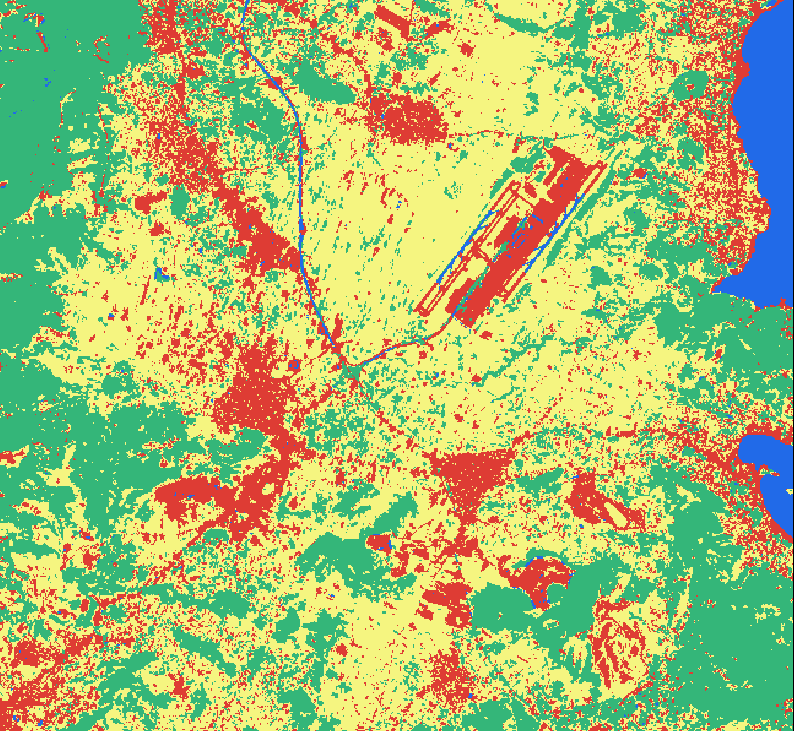}
	\end{minipage}
	\begin{minipage}[t]{0.32\textwidth}
		\includegraphics[width=\textwidth]{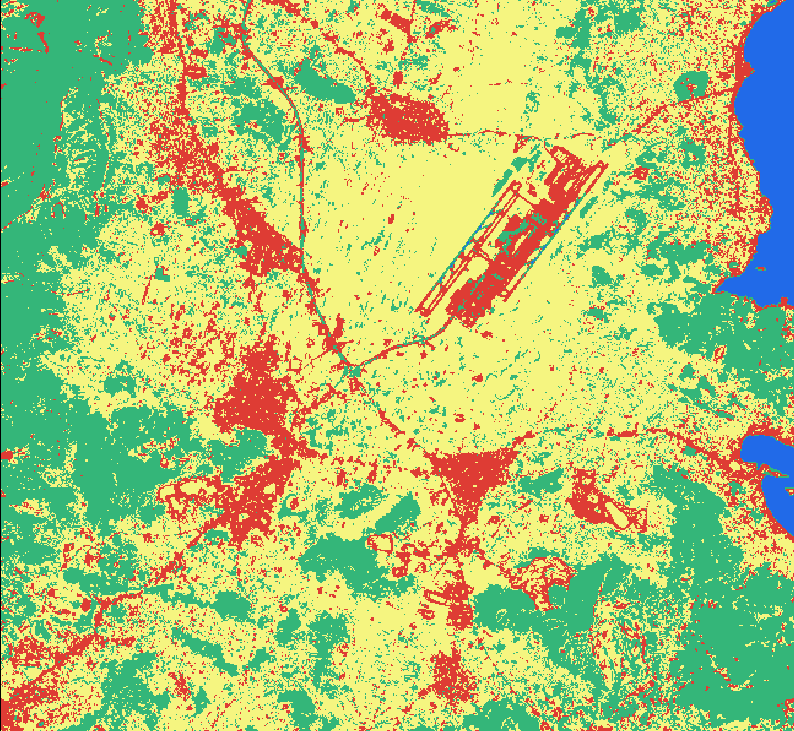}
	\end{minipage}
	\caption{Image characterization using (a) single barycenters (left) ($82.94\%)$, (b) two local barycenters for each environment type (middle) ($87.87\%$) and (c) multiple barycenters most efficient model (right) ($91.54\%$)}\label{fig-3.5}
\end{figure}

In general, a good choice for the number of sub-clusters in categories that present high heterogeneity should balance between accuracy improvement and the possible effect of over-parameterization. In Table \ref{tab-3.4} is illustrated the accuracy of the classification procedure for different choices of sub-clusters. Clearly, an improvement of about 10\% in total accuracy is succeeded comparing to the accuracy succeeded with single barycenters, justifying the further information allocation to more local centroids.

\section{Conclusions}

In this work clustering schemes for uncertain and structured data relying on the concept of Wasserstein barycenters are considered accompanied by appropriate clustering criteria based on the intrinsic geometry of the Wasserstein space which is selected to play the role of the feature space for the data in order to be better studied. The proposed geodesic criterion provides excellent performance in assessing both the quality of clustering and a measure of suitability for each observation to the cluster that is assigned to. The performance of both clustering scheme and the proposed index is illustrated in a simulation study and two real-world applications from the fields of econometrics and remote sensing. 

\bibliographystyle{apacite}
\bibliography{references}

\end{document}